\documentclass[letterpaper, 10 pt, conference]{ieeeconf}
\IEEEoverridecommandlockouts                              
\usepackage{color}
\usepackage{graphicx}
\usepackage[nameinlink,capitalise]{cleveref}
\crefname{line}{line}{lines}
\crefname{figure}{Fig.}{Figs.}
\Crefname{figure}{Fig.}{Figs.}
\crefname{equation}{Eq.}{Eqs.}
\Crefname{equation}{Eq.}{Eqs.}
\crefname{section}{Sec.}{Secs.}
\Crefname{section}{Sec.}{Secs.}
\crefname{definition}{Def.}{Defs.}
\Crefname{definition}{Def.}{Defs.}
\crefname{algorithm}{Alg.}{Algs.}
\Crefname{algorithm}{Alg.}{Algs.}

\begin{document}
\graphicspath{{./images/}}

\title{Encoding Reusable Multi-Robot Planning Strategies as Abstract Hypergraphs}

\author{Khen Elimelech$^{*,1}$, James Motes$^{*,3}$, Marco Morales$^{3,4}$, Nancy M. Amato$^{3}$,\\  Moshe Y. Vardi$^{1}$, and Lydia E. Kavraki$^{1,2}$
\thanks{$^*$~Equal contribution.}
\thanks{$^{1}$~The Department of Computer Science, Rice University,  Houston,\linebreak TX 77005, USA. {\tt\footnotesize \{elimelech,vardi,kavraki\}@rice.edu}}
\thanks{$^2$~Ken Kennedy Institute, Rice University, Houston, TX 77005, USA.}
\thanks{$^{3}$~Parasol Lab, University of Illinois at Urbana-Champaign, Champaign, IL 61820, USA. {\tt\footnotesize \{jmotes2,moralesa,namato\}@illinois.edu}}
\thanks{$^{4}$ ITAM, Ciudad de México, 01080, México. }
}

\maketitle

\begin{abstract}
Multi-Robot Task Planning (MR-TP) is the search for a discrete-action plan a team of robots should take to complete a task. The complexity of such problems scales exponentially with the number of robots and task complexity, making them challenging for online solution. To accelerate MR-TP over a system's lifetime, this work looks at combining two recent advances: (i) Decomposable State Space Hypergraph (DaSH), a novel hypergraph-based framework to efficiently model and solve MR-TP problems; and \mbox{(ii) learning-by-abstraction,} a technique that enables automatic extraction of generalizable planning strategies from individual planning experiences for later reuse. Specifically, we wish to extend this strategy-learning technique, originally designed for single-robot planning, to benefit multi-robot planning using hypergraph-based MR-TP.
\end{abstract}

\section{Introduction}
In recent times, multi-robot systems (MRS) have been rapidly integrated into real-world applications, enhancing throughput and enabling the completion of complex tasks through inter-robot cooperation.
Enabling this cooperation requires efficient Multi-Robot Task Planning (MR-TP).\linebreak Unfortunately, solutions to such planning problems suffer from prohibitively large search spaces, which often scale exponentially with the number of robots and task complexity.

\begin{figure}[t]
    \centering
    \includegraphics[width=0.6\columnwidth]{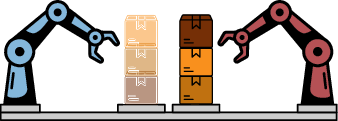}\\\vspace{-0.8\baselineskip}
    \textbf{a.}\dotfill\\\vspace{0.2\baselineskip}
    \includegraphics[width=0.65\columnwidth]{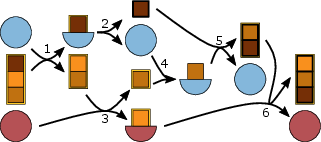}\\\vspace{-0.8\baselineskip}
    \textbf{b.}\dotfill\\\vspace{0.2\baselineskip}
    \includegraphics[width=0.6175\columnwidth]{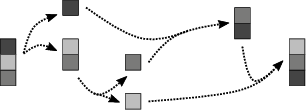}\\\vspace{-0.8\baselineskip}
    \textbf{c.}\dotfill\\\vspace{-0.4\baselineskip}
    \caption{
    \textbf{(a)}~Multi-Robot Task Planning problem: robots should re-stack the boxes from their current position on the right pedestal, into a desired position on the left pedestal; we may assume both robots can reach both pedastals.
    \textbf{(b)}~Solution hypergraph generated using DaSH, encoding a feasible action plan to complete the task: the blue robot first picks up the top box (represented by hyperarc 1); while it places it at the goal location, the red robot picks up the middle box (hyperarcs 2 and 3, respectively); then, the blue robot picks up the bottom box (hyperarc 4) and adds it to the goal stack (hyperarc 5); finally, the red robot places the box it is holding on top of the goal stack (hyperarc 6). 
    \textbf{(c)}~Abstract hypergraph, representing the generalizable solution strategy: all robot entities have been removed and explicit labels of the box entities have been stripped; the abstract hyperarcs (dashed) encode the progression of entity compositions.}
    \label{fig:orig-problem}
    \vspace{-20pt}
\end{figure}

Addressing this computational challenge, recent work introduced the Decomposable State Space Hypergraph (DaSH) framework~\cite{Motes23tro}, which succinctly models 
 and searches the MR-TP space using hypergraphs. 
Instead of searching for a trajectory in the composite state space, the hypergraph separates the problem into independent \emph{entities}, 
including the robots and manipulable objects, 
and captures the changes in (in)dependence of entity compositions due to action application throughout the solution process (see \cref{fig:orig-problem}). \linebreak
While this work focuses on task planning, DaSH is generally able to account for both task and motion constraints.
This model often produces a much more concise representation of the search space, resulting in faster planning times.

Nevertheless, even with this hypergraph-based representation, solving MR-TP problems remains challenging, especially when the number of robots or manipulable objects is large, or when the complexity of the multi-robot interactions required for a solution is high. Finding new techniques to improve the scalability of solution for MR-TP problems is thus an important standing challenge.

When facing a sequence of tasks, planning efficiency can potentially be improved throughout the MRS lifetime by exploiting and reusing successful planning experiences to avoid redundant computation on problems similar to those already solved.
In line with this idea, another recent work \cite{Elimelech22wafr,Elimelech22isrr,Elimelech23icra} established a novel approach for automatic extraction of generalizable planning strategies from individual successful experiences and their reuse to accelerate solution of new problems.
This approach was proved to be very effective~\cite{Elimelech24icra} for a single-robot planning.
This work seeks to extend the approach to support and benefit hypergraph-based MR-TP.
    \vspace{-2pt}

\section{Problem Definition}
    \vspace{-1pt}

Thus far, strategy extraction was only considered for single-robot planning, where the solution trajectory is depicted as a path in a conventional planning graph in the robot state space.
According to that approach, upon solution of a planning problem, one can identify a sequence of critical states from the solution trajectory, referred to as a Road~Map~(RM), which can then be abstracted (``lifted''), to yield an Abstract Road Map (ARM)---a sequence of abstract states representing a generalizable planning-strategy.\linebreak

\noindent Such an ARM can later be reconstructed (``grounded'') dynamically to match a new planning problem and then \textit{refined} into a complete solution trajectory, by independently resolving each abstract state-to-state transition as a sub-task.\linebreak In essence, we exploit (a generalization of) the solution structure extracted from the original problem to dynamically decompose and thus efficiently solve the new problem.

In this work, we shall adapt these concepts and processes, making them applicable to MR-TP and hypergraph-based solutions.

\section{Approach}
Next, we discuss how the solution abstraction and reuse procedures  can be adapted to hypergraph-based MR-TP.

\subsection{Step I: solution abstraction}
Since in \cite{Motes23tro}, MR-TP solutions are encoded as hypergraphs, rather than linear trajectories, an abstract MR-planning-strategy should be encoded as an \emph{Abstract~Hypergraph}~(AH), rather than an ARM.
Further, in single-robot planning, a robot was implicitly modeled in all states of the solution trajectory, as every action was always assumed to be preformed by one robot, sequentially.
Now, in \mbox{MR-TP}, robots are explicitly modeled in the solution hypergraph, as actions may be performed in parallel by different robots, or even require multiple robots to coordinate the performance of a single action.
Thus, we suggest now to also consider the notion of abstracting away all explicit robot entities from the hypergraph, resulting in an AH that implicitly models an \emph{abstract~robot} in every node.

\begin{figure}[t]
    \centering
    \begin{minipage}{0.49\columnwidth}
        \centering
        \includegraphics[width=\columnwidth]{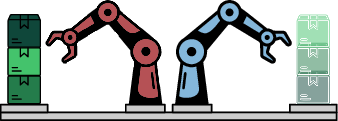}
    \end{minipage}\hfill
    \begin{minipage}{0.49\columnwidth}
        \centering
        \includegraphics[width=\columnwidth]{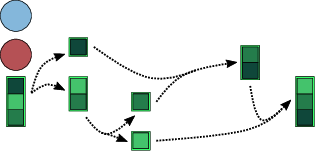}
    \end{minipage}
    \\[0\baselineskip]
    \makebox[0.5\linewidth][l]{\textbf{a.}\dotfill}\makebox[0.5\linewidth][r]{\textbf{b.}\dotfill}
    \\[0.2\baselineskip]
    \includegraphics[width=0.65\columnwidth]{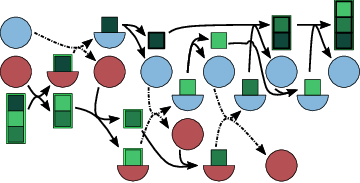}\\\vspace{-0.8\baselineskip}
    \textbf{c.}\dotfill\\

    \caption{\textbf{(a)} New planning problem: objects, locations of start and goal, robot position and reachability changed; moving boxes now requires handoffs. \textbf{(b)} Reconstructing the abstract hypergraph from \cref{fig:orig-problem}c to match this new problem. \textbf{(c)} Refining the reconstructed hypergraph into a complete solution: each abstract hyperarc represents a MR-TP sub-problem and is replaced with its solution hypergraph, generated using DaSH; handoff actions in the final solution are highlighted with dashed hyperarcs.}
    \label{fig:new-problem1}

    \vspace{-20pt}

\end{figure}
\begin{figure}[b]
    \centering
    \begin{minipage}{0.49\columnwidth}
        \centering
        \includegraphics[width=\columnwidth]{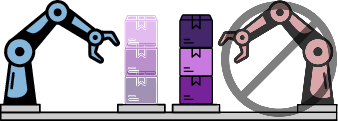}
    \end{minipage}\hfill
    \begin{minipage}{0.49\columnwidth}
        \centering
        \includegraphics[width=\columnwidth]{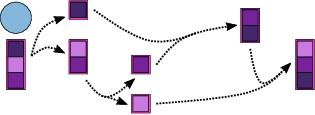}
    \end{minipage}
    \\[0.2\baselineskip]
    \makebox[0.5\linewidth][l]{\textbf{a.}\dotfill}\makebox[0.5\linewidth][r]{\textbf{b.}\dotfill}
    \\[0.2\baselineskip]
    \includegraphics[width=0.65\columnwidth]{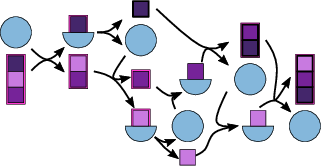}\\\vspace{-0.8\baselineskip}
    \textbf{c.}\dotfill\\
    
    \caption{\textbf{(a)} New planning problem: one robot is no longer operable. \textbf{(b)}~Reconstructing the abstract hypergraph from \cref{fig:orig-problem}c to match this new problem. \textbf{(c)} Refining the reconstructed hypergraph into a complete solution.}
    \label{fig:new-problem2}

\end{figure}

\paragraph*{Abstraction Example}
In \cref{fig:orig-problem}a, we see a pair of robots tasked to re-stack a box tower in a new order.
The hypergraph in \cref{fig:orig-problem}b models the transitions between entity compositions (nodes), as the robots manipulate the boxes, throughout the solution.
\cref{fig:orig-problem}c depicts the abstract hypergraph, after abstracting away all robot entities, selecting a subset of critical nodes (from the remaining nodes), and abstracting away all explicit object labels from those nodes. The order of the nodes is implicitly induced by the abstract hyperarcs.

\subsection{Step II: generalization and reuse}
To reuse the AH, we should, as before, reconstruct it, i.e., grounding some of the abstracted information with problem-specific details, and refining it, i.e., resolving each abstract hyperarc (replacing it with intermediate nodes and hyperarcs), to create a valid hyperpath, which connects all critical nodes in the proper order. 
In this process, the abstract robot that is implicitly modeled in each critical node may be grounded into any subset of robots (including the null set).\linebreak
By such, the AH formulation allows us to generalize a learned planning strategy to a variety of new problems, including problems with different robot-reachability constraints, numbers of robots, robot carrying capacity, etc..\linebreak
Let us look at a few examples of reusing the AH from \cref{fig:orig-problem}c.

\paragraph*{Reachability}
In Fig.~\ref{fig:new-problem1}a, the tower start and goal positions have changed; now, each robot can only reach one of the tower positions, requiring handoff actions to move boxes.\linebreak
First, we reconstruct the AH from \cref{fig:orig-problem}c to match this problem by grounding the object labels (as in previous work \cite{Elimelech22isrr}, this can be solved automatically through constraint satisfaction), and adding the relevant robot entities as initial nodes (\cref{fig:new-problem1}b).
We then refine the reconstructed hypergraph by treating each abstract hyperarc as a MR-TP sub-problem, which we solve using DaSH~\cite{Motes23tro}, with the new problem-specific robots and feasibility constraints (\cref{fig:new-problem1}c); \linebreak the solution hypergraphs are embedded in the reconstruced AH, replacing the abstract hyperarcs and adding robot entities to the critical entity-compositions (nodes) as needed.

\paragraph*{Number of Robots}
In Fig.~\ref{fig:new-problem2}a, one of robots is out of commission.
The matching process (Fig.~\ref{fig:new-problem2}b) is similar to the prior example, and, again, the refinement dynamically finds the necessary intermediate nodes and hyperarcs to reach the critical nodes in the specified order, with the single robot setting boxes down to the side until they are needed.

\section{Acknowledgements}

This work was supported in part by NSF-CCF-2336612, NSF-IIS-2123781, the IBM-Illinois Discovery Accelerator Institute, C-NICE at UIUC, and by Asociación Mexicana de Cultura AC.

\bibliographystyle{unsrt}
\bibliography{refs}
\end{document}